\documentclass[letterpaper]{article} 
\usepackage{aaai2026}  
\usepackage{times}  
\usepackage{helvet}  
\usepackage{courier}  
\usepackage[hyphens]{url}  
\usepackage{graphicx} 
\usepackage{natbib}  
\usepackage{caption}
\usepackage{amsmath,amssymb}
\usepackage{soul,color}
\usepackage{subcaption}
\usepackage[ruled,vlined]{algorithm2e}
\usepackage{booktabs} 
\usepackage{enumitem}
\usepackage{newfloat}
\usepackage{listings}
\DeclareCaptionStyle{ruled}{labelfont=normalfont,labelsep=colon,strut=off} 
\makeatletter\def\@listi{\leftmargin\leftmargini \topsep .5em \parsep .5em \itemsep .5em}
\def\@listii{\leftmargin\leftmarginii \labelwidth\leftmarginii \advance\labelwidth-\labelsep \topsep .4em \parsep .4em \itemsep .4em}
\def\@listiii{\leftmargin\leftmarginiii \labelwidth\leftmarginiii \advance\labelwidth-\labelsep \topsep .4em \parsep .4em \itemsep .4em}\makeatother

\newcounter{checksubsection}
\newcounter{checkitem}[checksubsection]

\newcommand{\checksubsection}[1]{%
  \refstepcounter{checksubsection}%
  \paragraph{\arabic{checksubsection}. #1}%
  \setcounter{checkitem}{0}%
}

\newcommand{\checkitem}{%
  \refstepcounter{checkitem}%
  \item[\arabic{checksubsection}.\arabic{checkitem}.]%
}
\newcommand{\question}[2]{\normalcolor\checkitem #1 #2 \color{blue}}
\newcommand{\ifyespoints}[1]{\makebox[0pt][l]{\hspace{-15pt}\normalcolor #1}}

\title{Fairness-Aware Few-Shot Learning for Audio-Visual Stress Detection}
\author {
    Anushka Sanjay Shelke\textsuperscript{\rm}\thanks{Corresponding author: anushka19@iiserb.ac.in},
    Aditya Sneh\textsuperscript{\rm },
    Arya Adyasha\textsuperscript{\rm },
    Haroon R. Lone\textsuperscript{\rm }
}
\affiliations {
    \textsuperscript{\rm} Indian Institute of Science Education and Research Bhopal, India\\
    \{anushka19, adityas19, arya20, haroon\}@iiserb.ac.in
}

\usepackage{bibentry}
\usepackage{amssymb}
\usepackage{adjustbox}
\usepackage{array}

\begin{document}

\maketitle

\begin{abstract}

Fairness in AI-driven stress detection is critical for equitable mental healthcare, yet existing models frequently exhibit gender bias, particularly in data-scarce scenarios. To address this, we propose \textbf{FairM2S}, a fairness-aware meta-learning framework for stress detection leveraging audio-visual data. FairM2S integrates Equalized Odds constraints during both meta-training and adaptation phases, employing adversarial gradient masking and fairness-constrained meta-updates to effectively mitigate bias. Evaluated against five state-of-the-art baselines, FairM2S achieves 78.1\% accuracy while reducing the Equal Opportunity to 0.06, demonstrating substantial fairness gains. We also release SAVSD, a smartphone-captured dataset with gender annotations, designed to support fairness research in low-resource, real-world contexts. Together, these contributions position FairM2S as a state-of-the-art approach for equitable and scalable few-shot stress detection in mental health AI. We release our dataset and FairM2S publicly with this paper at \url{https://tinyurl.com/48zzvesh}.

\end{abstract}

\section{Introduction}

Fairness in artificial intelligence (AI) and machine learning (ML) aims to build equitable systems that avoid bias and provide inclusive solutions for global issues like mental health \cite{qin2023towards,ferrara2024fairness}. Without it, AI in healthcare risks unfair outcomes for underrepresented groups, harming diagnosis, treatment, and care quality—especially in sensitive areas like mental health \cite{chinta2024ai}. For example, depression detection models trained mostly on male data suffer a 13–15\% accuracy drop on female samples, increasing misdiagnosis and inadequate care risk \cite{bailey2021gender}. Therefore, addressing fairness is critical for trustworthy, inclusive healthcare AI.

Amid growing concerns about fairness in mental health, stress detection is also affected. Stress is a major global health problem, with 31\% of adults naming it their top health concern and 62\% saying it disrupts daily life. It affects genders unequally, with 66\% of women and 58\% of men\footnote{\url{https://www.ipsos.com/en-us/ipsos-world-mental-health-day-report} \label{fnote-web1}}. To address this, audio-visual signals like facial expressions and voice traits offer a promising, easy way to detect stress using smartphones and webcam \cite{ghose2024integrating}, allowing real-time monitoring in natural settings. However, ensuring AI systems assess stress fairly across genders remains a major challenge.

Gender bias in machine learning models stems from imbalanced datasets, missing demographic data, and optimization focused on accuracy over fairness \cite{mehrabi2021survey}. For example, \citet{yang2024deconstructing} showed that speech-based models inherit demographic disparities, with gender differences in voice and symptom expression causing uneven results. Physiological factors like hormone-driven changes in heart rate and cortisol also lead to poorer performance in female data \cite{calderon2024gender}. Biases in data and algorithms worsen inequalities in healthcare \cite{mehrabi2021survey}. This is critical because women are more likely to experience stress than men, with studies showing significantly higher stress levels impacting their daily lives\footref{fnote-web1}.  These disparities emphasize the need for fairness-focused stress detection systems.

Moreover, data limitations worsen gender bias in mental health datasets, as many public datasets lack comprehensive gender labels and about 27.5\% of participants withhold demographic information due to privacy concerns and discrimination experiences \cite{hosseini2025faces,nong2022discrimination}. High costs of collecting large, diverse datasets lead to small, skewed samples. Conventional fairness methods like oversampling, reweighting, or adversarial debiasing often fail or cause notable accuracy loss; post-hoc debiasing can reduce accuracy by over 6\% \cite{zhang2018mitigating}. 

Despite advances in multimodal emotion and stress detection, gender bias persists in state-of-the-art models. Gender-specific CNN and BiLSTM classifiers show accuracy gaps up to 9\% favoring males \cite{zhang2023deep}. Facial expression recognition performs worse on female faces, with males detected more accurately for emotions like “Surprise” \cite{hosseini2025faces}. Speech-based emotion recognition similarly skews: female voices are classified better for positive emotions, males for negative \cite{yang2024deconstructing}. These disparities span CNNs, Vision Transformers, and self-supervised encoders, indicating systemic gender bias driven by data imbalance, label skew, and modality sensitivity \cite{hosseini2025faces,green2025gender,nong2022discrimination}.

Many fairness-aware few-shot learning methods apply fairness constraints only during meta-update or evaluation, allowing bias to persist during early adaptation—especially problematic for small, imbalanced multimodal stress detection datasets \cite{zhao2020fair}. These models also often fail to capture temporal context or maintain stable decision boundaries with noisy, limited data, leading to overconfident and unfair predictions \cite{luo2025fairness, mienye2024recurrent}. Additionally, current approaches frequently overlook biased updates from conflicting gradients between demographic groups, hindering fairness during generalization to unseen tasks \cite{zeng2024fairness}. This raises a central question:
\textbf{can fairness-aware few-shot learning models achieve high accuracy and fair stress detection by reducing bias throughout learning, even with limited data?}

To answer this question, we propose \textbf{FairM2S} — {\underline{Fair}ness-aware \underline{M}eta-learning for \underline{M}ultimodal \underline{S}tress Detection}, a fairness-aware meta-learning framework for few-shot multimodal stress detection using audio-visual data. By applying Equalized Odds constraints through adversarial gradient masking, gradient projection and fairness-constrained updates during meta-training and adaptation, it reduces gender bias. FairM2S achieves 78.1\% accuracy and an Equal Opportunity (Eopp) of 0.062, outperforming five baseline models in both performance and fairness. To support fairness research in low-resource settings, we introduce \textbf{Stress Audio-Visual Speech Dataset (SAVSD)}, a smartphone-collected dataset with gender annotations. Together, these position FairM2S as a scalable, equitable solution for stress detection. Our key contributions are:
\begin{itemize} [itemsep=0em,parsep=0em]
    \item \textbf{FairM2S framework:} A fairness-aware meta-learning framework mitigating gender bias in few-shot multimodal stress detection.
    \item \textbf{SAVSD dataset:} A low-cost, smartphone-based multimodal dataset\footnote{\url{https://tinyurl.com/48zzvesh}} with gender annotations for fairness evaluation in resource-limited settings.

    \item \textbf{Benchmark evaluation:} A scalable, non-intrusive stress detection benchmark demonstrating superior accuracy and fairness across datasets and modalities.
\end{itemize}

\section{Related Work}
\subsection{Gender Fairness}

 Gender fairness in mental health AI has advanced significantly in depression, anxiety, and emotion recognition, yet it remains largely underexplored in stress detection \cite{yang2024deconstructing,hemakom2023ecg}. Techniques like adversarial training and domain adaptation have proven effective at reducing gender bias on large datasets; for example, improvements of over 13 percentage points in depression and PTSD detection were achieved using 73.6 hours of E-DAIC data, and gender performance gaps were reduced below 2\% on 324 hours of MSP-Podcast audio with minimal accuracy loss \cite{kim2025domain,lin2025emo}. However, such large, diverse datasets are rare in stress detection, which typically involves smaller, less comprehensive samples, limiting the direct applicability of these methods to stress-related tasks. Nonetheless, mobile health studies show promise in improving fairness even with small cohorts \cite{park2022fairness}. Likewise, Jiang et al. \cite{jiang2024evaluating} studied 132 participants across modalities and showed that post-training threshold optimization improves fairness metrics with only a slight drop in F1-score. Early stress detection research reports substantial gender gaps without applying fairness mitigation techniques \cite{zuo2011cross,liapis2015stress}, underscoring the urgent need for fairness-aware methods specifically designed for stress detection in limited-data scenarios.

\subsection{Fair Few Shot Learning}

Recent work addresses gender bias under data scarcity through fairness-aware meta- and few-shot learning. While Model-Agnostic Meta-Learning (MAML) enables rapid adaptation, it overlooks fairness and often perpetuates bias \cite{finn2017model}. Building on this, Fair-MAML incorporates group fairness constraints during meta-training to achieve “K-shot fairness,” though it remains sensitive to label sparsity and distribution shifts \cite{slack2020fairness}. Decision Boundary Covariance methods penalize correlations between sensitive attributes and predictions in few-shot tasks but require complete demographic labels \cite{zhao2020fair}. Similarly, FORML improves worst-group metrics by dynamically reweighting samples, yet struggles with imbalanced and multimodal data \cite{yan2202forml}. Nash Bargaining Meta-Learning tackles subgroup conflicts via a two-stage game, boosting fairness up to 67\%, but is limited by static group definitions and computational costs \cite{zeng2024fairness}. Lagrange duality approaches jointly optimize privacy and fairness but depend on accurate annotations and careful parameter tuning \cite{wang2025privacy}. Finally, FEAST transfers fairness knowledge across tasks using auxiliary data but assumes high-quality, balanced group labels, which are rare in mental health contexts \cite{wang2023fair}. In contrast, FairM2S enforces fairness with differentiable Equalized Odds loss during inner and outer-loop updates using adversarial masking and gradient projection, overcoming prior limits to enable practical mental health deployment in low-resource settings.

\subsection{Existing Datasets}

Multimodal datasets have advanced stress detection but are limited by small samples, restricted diversity, access barriers, costly equipment, and indirect emotion-to-stress mappings \cite{livingstone2018ryerson,chaptoukaev2023stressid,zhou2023investigating}. Emotion datasets like RAVDESS, IEMOCAP, and CREMA-D, with small cohorts (24, 10, 91), use anger, fear, and disgust as stress proxies but lack cultural diversity, rely on actors, and sometimes require registration, limiting accessibility and naturalism \cite{livingstone2018ryerson,busso2008iemocap,cao2014crema}. Stress datasets like StressID, ForDigitStress, MuSE, WESAD, and BESST (65, 40, 28, 15, and 79 participants respectively) offer multimodal physiological, audio, and visual data under controlled stress induction (e.g., TSST). However, they are mostly Western-centric, small-scale, often lack complete demographics like gender, and rely on costly clinical-grade sensors such as ECG and chest straps, limiting their use in resource-constrained settings.
\cite{chaptoukaev2023stressid,heimerl2024fordigitstress,jaiswal2020muse,schmidt2018introducing,pevsan2024besst} Naturalistic datasets like SWELL-KW and K-EmoCon (25 and 32 participants) offer real-world recordings from workplaces and social debates, improving ecological validity but suffer from small cohorts, limited demographics, and restricted access
\cite{koldijk2014swell,park2020k}. Cross-corpus studies show physiological stress detection models (ECG, EDA) capture general high-arousal, not stress-specific, states. Similar generalization to non-stress high-arousal emotions highlights the limits of relying solely on indirect emotion-to-stress mappings or physiological markers \cite{zhou2023investigating}.We introduce \textbf{SAVSD}, an audio-visual dataset of 128 East Asian participants recorded via low-cost smartphones during TSST, annotated with DASS-21 stress scores and gender, for scalable, fairness-aware stress detection in low-resource settings.

\section{Problem Formulation}
We address the challenge of fair few-shot learning for binary stress classification from multimodal sequential data. Let \(\mathcal{D} = \{(\mathbf{x}_i, y_i, g_i)\}_{i=1}^N\) denote the dataset, where \(N\) is the total number of participants, and each participant indexed by \(i \in \{1, \dots, N\}\) has a fused audio-video sequence \(\mathbf{x}_i \in \mathbb{R}^{T \times d}\), a binary stress label \(y_i \in \{0,1\}\), and a gender group label \(g_i \in \mathcal{G} = \{\text{Male}, \text{Female}\}\), with \(T\) as the sequence length and \(d\) is the number of features per time step. To tackle both accuracy and fairness, our approach builds upon episodic meta-learning, which has demonstrated superior few-shot learning (FSL) performance \cite{gharoun2024meta,finn2017model}. In FSL, the dataset \(\mathcal{D}\) is split into disjoint meta-training \(\mathcal{D}_{\mathrm{train}}\) and meta-testing \(\mathcal{D}_{\mathrm{test}}\) sets, ensuring no participant overlap (\(\mathcal{D}_{\mathrm{train}} \cap \mathcal{D}_{\mathrm{test}} = \emptyset\)). During training, the model learns from tasks $\mathcal{T}_j$, each representing a small few-shot problem drawn from the task distribution $\mathcal{T}$. Here, task represents a small few-shot classification problem for detecting stress on new participants. Each task contains a support set \(\mathcal{S}_j = \{(\mathbf{x}_i, y_i, g_i)\}_{i=1}^K\) of \(K\) labeled examples used for adaptation, and a query set \(\mathcal{Q}_j = \{(\mathbf{x}_i, y_i, g_i)\}_{i=1}^Q\) of \(Q\) examples used for evaluation during training. The meta-testing set \(\mathcal{D}_{\mathrm{test}}\) is reserved for evaluating the model’s performance on unseen participants. The objective is to learn a classifier \(f_\theta: \mathbb{R}^{T \times d} \to [0,1]\), parameterized by \(\theta\), that estimates stress probability.

Fairness is ensured via the Equalized Odds criterion, which enforces equal true and false positive rates (TPR and FPR) across gender groups \(g \in \mathcal{G}\) while capturing both error types—making it better suited for stress detection, where both false alarms and missed detections are harmful~\cite{hardt2016equality}. The Equalized Odds gap is defined as \(\mathrm{Eodd} = \frac{1}{2} \big(|\mathrm{TPR}_{g_1} - \mathrm{TPR}_{g_2}| + |\mathrm{FPR}_{g_1} - \mathrm{FPR}_{g_2}|\big)\), where \(g_1, g_2\) denote the groups, Male and Female. To balance accuracy and fairness, we minimize the expected sum of classification, fairness, and regularization losses across tasks:

\(\min_{\theta} \mathbb{E}_{\mathcal{T}_j \sim \mathcal{T}} \big[ \mathcal{L}_{\mathrm{cls}}(f_\theta, \mathcal{Q}_j) + \gamma \, \mathcal{L}_{\mathrm{Eodd}}(f_\theta, \mathcal{Q}_j) + \alpha \, \mathcal{L}_{\mathrm{margin}}(f_\theta, \mathcal{S}_j) + \beta \, \mathcal{L}_{\mathrm{smooth}}(f_\theta, \mathcal{S}_j) \big]\)

Here, \(\mathbb{E}_{\mathcal{T}_j \sim \mathcal{T}}[\cdot]\) denotes expectation over tasks, reflecting average few-shot performance. \(\mathcal{L}_{\mathrm{cls}}\) is binary cross-entropy loss on query sets for accuracy; \(\mathcal{L}_{\mathrm{Eodd}}\) penalizes fairness gaps; \(\mathcal{L}_{\mathrm{margin}}\) enforces class margin on support sets; and \(\mathcal{L}_{\mathrm{smooth}}\) applies label smoothing to reduce overconfidence. Hyperparameters \(\gamma, \alpha, \beta\) balance these terms.

\section{Proposed Method}
This section presents our framework \textbf{FairM2S} outlining the model backbone, fairness-aware loss components, and gradient-based optimization strategies used in both inner-loop adaptation and outer-loop meta-updates to achieve balanced and fair learning (see Algorithm~\ref{alg:fairm2s}).

\subsection{Backbone Architecture}

Existing detection models often use unidirectional recurrent networks or simple feature aggregation, limiting their ability to capture temporal context before and after each moment and missing subtle progressive stress signals \cite{mienye2024recurrent}. FairM2S addresses this by using a Bidirectional LSTM (BiLSTM) to process sequences forward and backward, capturing rich context at each time step \cite{zhao2021multi}. A GRU layer then refines these features, emphasizing relevant signals and reducing noise \cite{aynali2020noise}. Outputs are pooled via global average pooling into a fixed-length embedding, passed through a fully connected layer with dropout for robustness, and a sigmoid layer outputs stress probability. This backbone effectively learns temporal patterns from fused audio-video data.

\subsection{Fair Inner Loop Meta-Adaptation}

 In gradient-based meta-learning algorithms, the inner loop refers to the rapid, task-specific adaptation phase where the model updates its parameters \(\theta\) to task-adapted parameters \(\theta_j'\) for a given task \(\mathcal{T}_j\). This adaptation is performed using the support set \(\mathcal{S}_j\), allowing quick adaptation of \(\theta\) by minimizing a task-specific loss. Traditional methods such as MAML~\cite{finn2017model}, ANIL~\cite{yuksel2024first}, and MAML-TSC~\cite{wang2024maml} primarily optimize for accuracy using the classification loss, often overlooking fairness. This can result in biased models, especially when the support set is small and imbalanced across gender groups. In such cases, models may overfit to spurious correlations (misleading correlations) or dominant group features, allowing bias to propagate during meta-training and ultimately yield unfair predictions~\cite{mehrabi2021survey, ochal2021sensitive}.

To overcome these issues, \textbf{FairM2S} explicitly incorporates fairness and regularization terms into the inner-loop loss, defined as \(\mathcal{L}_{\mathrm{inner}} = \mathcal{L}_{\mathrm{cls}} + \gamma \mathcal{L}_{\mathrm{Eodd}} + \alpha \mathcal{L}_{\mathrm{margin}} + \beta \mathcal{L}_{\mathrm{smooth}}\), where \(\gamma\), \(\alpha\), and \(\beta\) are hyperparameters balancing accuracy, fairness, and regularization. This early-stage integration avoids the limitations of post-hoc fairness correction~\cite{yang2023algorithmic}. The \textbf{Equalized Odds loss} \(\mathcal{L}_{\mathrm{Eodd}}\) specifically addresses disparities that arise when support sets are small and imbalanced across gender groups. In such settings, gradient updates may favor the dominant group, leading to unfair model behavior. \(\mathcal{L}_{\mathrm{Eodd}}\) mitigates this by minimizing the gap between true and false positive rates across groups~\cite{xu2022algorithmic}. In our setup, positive samples represent stressed states (\(y_i = 1\)) and negative samples denote non-stressed states (\(y_i = 0\)). For a group \(g \in \mathcal{G}\), let \(P_g = \{i : g_i = g, y_i = 1\}\) and \(N_g = \{i : g_i = g, y_i = 0\}\) denote the sets of positive and negative samples for group \(g\), respectively. Group-wise metrics are computed as \(\mathrm{TPR}_g = \frac{1}{|P_g|} \sum_{i \in P_g} f_\theta(\mathbf{x}_i)\) and \(\mathrm{FPR}_g = \frac{1}{|N_g|} \sum_{i \in N_g} f_\theta(\mathbf{x}_i)\). We apply a fairness loss \(\mathcal{L}_{\mathrm{Eodd}}\) on these quantities to encourage equalized error rates during adaptation.

To address the accuracy–fairness trade-off in ambiguous few‑shot tasks—where small, imbalanced support sets can lead to unreliable decision boundaries—FairM2S incorporates a \textbf{margin loss} \(\mathcal{L}_{\mathrm{margin}}\), inspired by adaptive margin strategies in few‑shot literature \cite{luo2025fairness} to encourage a stable decision boundary despite limited data. Let \(S_+ = \{i : y_i = 1\}\) and \(S_- = \{i : y_i = 0\}\) denote the positive (stressed) and negative (non‑stressed) support indices, respectively. We compute mean logits \(\bar{z}_+ = \frac{1}{|S_+|} \sum_{i \in S_+} f_\theta(\mathbf{x}_i)\) and \(\bar{z}_- = \frac{1}{|S_-|} \sum_{i \in S_-} f_\theta(\mathbf{x}_i)\). The margin loss is then defined as \(\mathcal{L}_{\mathrm{margin}} = \max\bigl(0,\; m - (\bar{z}_+ - \bar{z}_-)\bigr)\), where \(m > 0\) is a margin hyperparameter. By enforcing a minimum gap between positive and negative class logits, this term strengthens class separability, guides learning away from spurious or dominant-group features, and thereby indirectly contributes to fairness while improving robustness~\cite{lee2019meta}. Because few-shot support sets are small and prone to noise or imbalance, models can become overconfident and overfit to individual labels, which may exacerbate fairness issues. To mitigate overconfidence on limited data, we incorporate \textbf{label smoothing}. The loss \(\mathcal{L}_{\mathrm{smooth}}\) softens binary targets (e.g., converting 1 to 0.9), helping the model avoid overfitting and improving its generalization capacity—an often overlooked challenge in few-shot settings~\cite{gao2022label}.

Finally, the model parameters are updated using gradient descent on the combined inner loss: \(\theta_j' = \theta - \eta \nabla_\theta \mathcal{L}_{\mathrm{inner}}(f_\theta, \mathcal{S}_j)\), where \(\eta\) is the inner-loop learning rate. This update ensures that each task-specific adaptation step jointly optimizes for fairness and accuracy, enabling FairM2S to produce robust and equitable stress classifiers across diverse participant groups.

\subsection{Fair Outer Loop Meta Adaptation}

After the inner loop adapts \(\theta\) to task \(\mathcal{T}_j\), yielding \(\phi_j = \theta_j'\), the outer loop aggregates information across tasks to refine the shared initialization \(\theta\). Traditional methods often defer fairness enforcement to final optimization or evaluation, which risks bias into meta-parameters. FairM2S instead integrates fairness directly into the meta-gradient computation, preventing biased updates and improving fairness during generalization to unseen tasks~\cite{zeng2024fairness}.

\begin{algorithm}[]
\caption{FairM2S}
\label{alg:fairm2s}
\LinesNumbered
\KwIn{Dataset $\mathcal{D}$, task distribution $\mathcal{T}$, group set $\mathcal{G}$, meta-parameters $\theta$, inner-loop learning rates $\eta$, outer-loop learning rates $\beta$, loss weights $\gamma$, $\alpha$, $\lambda$}
\KwOut{Trained meta-parameters $\theta$}

\textbf{Initialize} meta-parameters $\theta$ with random weights

\For{each meta-iteration}{
    Sample task batch $\{\mathcal{T}_j\} \sim \mathcal{T}$

    \For{each task $\mathcal{T}_j$ in batch}{
        Sample support set $\mathcal{S}_j$ and query set $\mathcal{Q}_j$

        \textbf{Inner Loop:} Initialize $\phi_j \gets \theta$ \\
        Update $\phi_j$ via gradient descent on loss: \\
        \quad $\mathcal{L}_{\mathrm{inner}} = \mathcal{L}_{\mathrm{cls}} + \gamma \mathcal{L}_{\mathrm{Eodd}} + \alpha \mathcal{L}_{\mathrm{margin}} + \lambda \mathcal{L}_{\mathrm{smooth}}$ using $\mathcal{S}_j$

        \textbf{Outer Loop:} Compute group-wise gradients $g^{(g)} = \nabla_\theta \mathcal{L}_q^{(g)}(\phi_j)$ and full gradient $g = \nabla_\theta \mathcal{L}_q(\phi_j)$ \\
        Apply adversarial mask: $g^{\mathrm{adv}} = A(g) \odot g$ \\
        Project gradient: $g^{\mathrm{fair}} = \text{Proj}(g^{\mathrm{adv}}, \{g^{(g)}\})$
    }

    \textbf{Meta-update:} \\
    \quad $\theta \leftarrow \theta - \beta \cdot \frac{1}{|\{\mathcal{T}_j\}|} \sum_j g_j^{\mathrm{fair}}$
}
\end{algorithm}

A key challenge in fairness-aware meta-learning arises from imbalanced query sets, where gradients disproportionately reflect the dominant group, leading to biased meta-updates. To address this, FairM2S first computes \textbf{Group-Specific Gradients}. For each task \(\mathcal{T}_j\), the query loss is \(\mathcal{L}_q(\phi_j, \mathcal{Q}_j) = \frac{1}{Q} \sum_{i=1}^Q \ell(f_{\phi_j}(\mathbf{x}_i), y_i)\), where \(\ell\) is binary cross-entropy loss. The query set is partitioned by sensitive group \(g_i \in \mathcal{G}\), yielding subsets \(\mathcal{Q}_g = \{i : g_i = g\}\), with group-wise losses defined as \(\mathcal{L}_q^{(g)}(\phi_j) = \frac{1}{|\mathcal{Q}_g|} \sum_{i \in \mathcal{Q}_g} \ell(f_{\phi_j}(\mathbf{x}_i), y_i)\). The corresponding gradients \(g^{(g)} = \nabla_\theta \mathcal{L}_q^{(g)}(\phi_j)\) indicate how \(\theta\) would be updated if optimized solely for group \(g\). Misalignment among gradients reveals inter-group conflicts, signaling potential bias. Detecting and analyzing such conflicts is central to promoting fairness and builds on prior multi-task and fairness-aware learning strategies~\cite{zeng2024fairness, malik2025faalgrad}.

Group-specific gradients help identify inter-group conflicts \& reveal bias directions, they do not actively constrain biased updates \cite{selialia2024mitigating}. To address this, our framework used an \textbf{Adversarial Gradient Masking} mechanism. The overall meta-gradient is \(g = \nabla_\theta \mathcal{L}_q(\phi_j, \mathcal{Q}_j)\). A lightweight adversarial network \(A(\cdot)\) receives \(g\) and outputs a mask \(m = A(g)\), where \(m \in (-1,1)\) via \(\tanh\) activation. The masked gradient is then computed as \(g^{\mathrm{adv}} = m \odot g\), where \(\odot\) denotes the Hadamard (element-wise) product \cite{luo2025fairness}. This adversary is trained jointly with the meta-learner to downweight gradient components associated with unfair updates, thereby filtering out bias-amplifying directions. Building on adversarial fairness and gradient filtering~\cite{luo2025fairness}, this approach complements group-gradient analysis with dynamic, data-driven fairness during meta-optimization.


To  neutralize update directions that may amplify group disparities, FairM2S employs a \textbf{Fairness-Constrained Gradient Projection} step during meta-optimization. For two groups, the disparity direction is defined as \(d = g^{(0)} - g^{(1)}\), representing the axis along which updates would increase inter-group performance differences. To prevent this, the adversarially masked gradient is projected orthogonally to \(d\), yielding a fairness-aware update: \(g^{\mathrm{fair}} = g^{\mathrm{adv}} - \frac{\langle g^{\mathrm{adv}}, d \rangle}{\|d\|^2 + \varepsilon} d\), where \(\langle \cdot, \cdot \rangle\) denotes the dot product and \(\varepsilon\) is a small constant for numerical stability \cite{yu2020gradient, hsieh2024careful, zeng2024fairness}. For more than two groups (\(M > 2\)), the disparity direction generalizes to \(d = \sum_{i < j} (g^{(i)} - g^{(j)})\), capturing the aggregate inter-group disagreement. The final meta-parameter update is then performed as \(\theta \leftarrow \theta - \beta g^{\mathrm{fair}}\), with \(\beta\) denoting the meta learning rate. This projection complements adversarial masking by neutralizing biased update directions, ensuring that the learned initialization supports equitable adaptation across groups. The learned meta-parameters \(\theta\) are the final trained model initialization obtained after meta-training, which can be quickly adapted to new tasks for accurate and fair stress classification.

\section{Experiments}

\subsection{Datasets} 

We used three datasets in our analysis: our collected SAVSD, and two public ones—StressID \cite{chaptoukaev2023stressid} and Audio Video Dataset (AVD) \cite{sahu2024unveiling}. Table~\ref{tab:datasets}summarizes their demographics.

\paragraph{SAVSD:}The dataset was collected at an \textit{anonymous} institute, with ethical approval obtained from the institutional review board. It includes audio-visual recordings from 128 student participants (94 males, 34 females) performing a standardized stress-inducing speech activity, a Trier Social Stress Test~\cite{kirschbaum1993trier}. Recordings were captured using a Redmi 9A smartphone mounted on a tripod to ensure consistency and accessibility. Each participant was informed their speech would be evaluated, assigned a random topic validated by clinical psychiatrists, given 20 seconds to prepare, and spoke for two minutes. Post-speech, participants completed a 7-item DASS-21 \cite{lovibond1995manual} based self-report questionnaire on a 4-point Likert scale (0--3). Summed scores classified participants as ``Stressed'' (66 total: 45 males, 21 females) or ``Non-Stressed'' (62 total: 49 males, 13 females). The dataset includes short audio-visual clips with demographic and stress labels. The smartphone-based, low-cost protocol supports reproducible in resource-constrained settings.

\paragraph{StressID:}  It is a widely recognized public benchmark dataset for stress recognition. Of the 65 original participants, our study includes video recordings from 53 available participants. We focused on the ``Speaking'' segment—a \textit{Social Evaluative Task (SET)} where participants described their strengths and weaknesses in a job interview-like setting to induce stress. Each segment lasted approximately one minute. Participants were labeled as ``Stressed'' or ``Non-Stressed'' based on self-reported and physiological indicators provided in the dataset.

\paragraph{AVD:}
This dataset includes audio-visual recordings of 101 participants, each delivering a stress-inducing speech for at least 2.5 minutes. Immediately after the speech, participants completed a brief two-item questionnaire:
(i) \textit{I felt extreme nervousness in the last activity.} (1: Strongly Disagree to 5: Strongly Agree)
(ii) \textit{During the last activity, how did you feel about yourself? }(1: Unhappy/Negative to 5: Happy/Positive)

To compute stress labels, we reversed the responses to question no. ii and calculated the mean score across both items. Participants with a mean $\geq 3$ were labeled as “Stressed,” otherwise “Not Stressed.” This resulted in 55 “Stressed” (35 males, 20 females) and 46 “Not Stressed” (31 males, 15 females).

\begin{table}[h]
\centering
\footnotesize
\begin{tabular}{lccl}
\toprule
Dataset & $\#$ & Age ($\mu\pm\sigma$) & Stress \textbf{/} Non-Stress (male, female) \\
\midrule
SAVSD    & 128 & 19.5$\pm$1.5 & 66(45, 21) {\bf/} 62(49, 13) \\
StressID &  53 & 29.0$\pm$8.5 & 37(28, 9) {\bf/} 16(12, 4) \\
AVD      & 101 & 21.1$\pm$3.3 & 55(35, 20) {\bf/} 46(31, 15) \\
\bottomrule
\end{tabular}
\caption{Participant demographics}
\label{tab:datasets}
\end{table}

\subsection{Data Preprocessing}

All SAVSD recordings were standardized to 120 seconds to ensure uniformity, addressing minor variations of approximately ±5 seconds. Recordings longer than 120 seconds were truncated, while shorter ones were zero-padded. Audio and video features were extracted using non-overlapping 1-second windows. For video, OpenFace \footnote{https://github.com/TadasBaltrusaitis/OpenFace} extracted 37 features per window, including facial action units (AUs), eye gaze, head pose, and facial landmarks. For audio, 30 features per second were extracted using Librosa \footnote{https://librosa.org/doc/latest/index.html}, including MFCC~${1-13}$, energy, F0, HNR, speaking rate, pause rate, spectral centroid, bandwidth, rolloff, contrast~${1-7}$, flatness, and ZCR—following standard speech processing practices~\cite{duvvuri2024unravelling, giannakakis2020automatic, giannakakis2022automatic, lu2024daily}. The StressID and AVD datasets were preprocessed similarly to ensure consistent sequence lengths and aligned feature representations across modalities for downstream analysis.

\subsection{Baseline Models}

We compare FairM2S against five baselines. \textbf{MAML} serves as our primary traditional baseline, offering task-specific adaptation without any fairness consideration, thereby highlighting the impact of fairness integration. Building on this, \textbf{Fair-MAML} incorporates fairness constraints into MAML’s optimization process to mitigate bias during few-shot learning. \textbf{FEAST} enhances fairness by leveraging auxiliary sets during meta-testing, enabling adaptation even under data imbalance across sensitive groups.\textbf{ Fair DBC }addresses bias by enforcing distributional balance constraints during adaptation, promoting equitable performance across groups. Finally, \textbf{Nash-Met} formulates fairness-aware meta-learning as a cooperative bargaining game using Nash Bargaining solutions to resolve hypergradient conflicts and ensure stable, fair convergence across groups. For a consistent and fair evaluation, all baselines are implemented using the same BiLSTM-GRU backbone architecture.

\subsection{Training Configuration and Evaluation Metrics}

All models were trained in a unified experimental setup to ensure fair comparison. We fixed the batch size at 32 and trained for 50 epochs using the Adam optimizer. The base inner-loop learning rate \(\eta\) and the meta-learning rate \(\beta\) were set to \(10^{-3}\) for all models. Few-shot tasks were configured as 2-way classification with 5 support examples (shots) per class for meta-adaptation and 15 query samples (\(Q=15\)) per task for evaluation. We evaluated all models using 1, 3, and 5 shots. \textbf{Hyperparameter tuning} was performed as follows: FairMAML and FairM2S tuned the fairness weight \(\gamma\) over \(\{0.01, 0.05, 0.1, 0.2, 0.5, 1.0\}\), with FairM2S additionally tuning the label smoothing factor \(\beta\) in \(\{0.05, 0.1, 0.2\}\) and the margin loss weight \(\alpha\) in \(\{0.1, 0.2, 0.3\}\). MAML tuned the meta learning rate \(\beta\) over \(\{0.0005, 0.001, 0.003, 0.005\}\). FairDBC searched the regularization coefficient \(\lambda_{\mathrm{dbc}}\) in \(\{0.01, 0.05, 0.1, 0.5, 1.0\}\) and used three inner-loop steps per meta-update. FEAST explored auxiliary set ratios \(\rho_{\mathrm{aux}} \in \{0.1, 0.2, 0.5\}\) and mutual information regularization weights \(\lambda_{\mathrm{mi}} \in \{0.05, 0.1, 0.5, 5.0\}\). NashNet varied the number of bargaining iterations in stage 1, \(n_{\mathrm{stage1}} \in \{10, 15, 20\}\). Each configuration run 5 times to evaluate the randomness and stability. These are baseline hyperparameters. \textit{All experiments were conducted on a system equipped with an NVIDIA RTX 4070 12GB GPU, Win 11, 64 GB RAM, and an Intel i9 12th Gen CPU.}

\textbf{Evaluation metrics:} We evaluate performance and fairness with Accuracy, Disparate Impact (DI), and Equal Opportunity (Eopp). Accuracy is the overall correct prediction rate. DI measures statistical parity by comparing positive prediction rates across genders; values near 1 indicate fairness. Eopp, the absolute difference in true positive rates between groups, ensures equal detection of stressed individuals. These metrics are strongly supported by recent fairness research in sensitive domains~\cite{pessach2022review,cheong2023towards,schmitz2022bias}.

\section{Results}
\renewcommand{\arraystretch}{1.25}
\begin{table*}[t]
\centering
\resizebox{\textwidth}{!}{
\begin{tabular}{c|c|ccc|ccc|ccc}
\toprule
\hline
\textbf{Datasets $\rightarrow$} & & \multicolumn{3}{c|}{\textbf{SAVSD}} & \multicolumn{3}{c|}{\textbf{StressID}} & \multicolumn{3}{c}{\textbf{AVD}} \\
\hline
\textbf{Model}& \textbf{Shots}  & \textbf{Accuracy} & \textbf{DI} & \textbf{Eopp} & \textbf{Accuracy} & \textbf{DI} & \textbf{Eopp} & \textbf{Accuracy} & \textbf{DI} & \textbf{Eopp} \\
\hline
& 1 & 0.544 ± 0.132 & \textbf{0.891 ± 0.099} & 0.129 ± 0.112 & 0.662 ± 0.095 & 0.821 ± 0.095 & 0.192 ± 0.109 & 0.600 ± 0.046 & 0.690 ± 0.168 & 0.180 ± 0.148 \\
\textbf{MAML} & 3 & 0.538 ± 0.119 & 0.752 ± 0.171 & 0.167 ± 0.115 & 0.662 ± 0.144 & 0.711 ± 0.256 & 0.200 ± 0.090 & 0.600 ± 0.068 & 0.642 ± 0.237 & 0.240 ± 0.167 \\
 & 5 & 0.553 ± 0.082 & 0.713 ± 0.111 & 0.204 ± 0.185 & 0.662 ± 0.130 & 0.843 ± 0.107 & 0.158 ± 0.116 & 0.600 ± 0.046 & 0.719 ± 0.175 & 0.180 ± 0.130 \\
\hline
 & 1 & 0.533 ± 0.088 & 0.810 ± 0.158 & 0.195 ± 0.085 & 0.675 ± 0.120 & \textbf{0.848 ± 0.128} & 0.258 ± 0.103 & 0.592 ± 0.074 & 0.520 ± 0.177 & 0.380 ± 0.109 \\
\textbf{Fair-MAML} & 3 & 0.523 ± 0.084 & 0.714 ± 0.086 & 0.252 ± 0.064 & 0.637 ± 0.161 & 0.799 ± 0.086 & 0.208 ± 0.147 & 0.571 ± 0.025 & 0.679 ± 0.201 & 0.280 ± 0.130 \\
 & 5 & 0.543 ± 0.081 & 0.773 ± 0.185 & 0.223 ± 0.031 & 0.687 ± 0.125 & 0.836 ± 0.101 & 0.183 ± 0.158 & 0.564 ± 0.029 & 0.675 ± 0.239 & 0.260 ± 0.219 \\
\hline
 & 1 & 0.538 ± 0.072 & 0.707 ± 0.215 & 0.190 ± 0.151 & 0.637 ± 0.149 & 0.843 ± 0.067 & 0.158 ± 0.116 & 0.571 ± 0.035 & 0.665 ± 0.206 & 0.320 ± 0.148 \\
\textbf{DBC F-MAML} & 3 & 0.564 ± 0.047 & \textbf{0.753 ± 0.169} & 0.133 ± 0.143 & 0.687 ± 0.139 & 0.776 ± 0.117 & 0.208 ± 0.155 & 0.564 ± 0.068 & 0.582 ± 0.171 & 0.259 ± 0.240 \\
 & 5 & 0.533 ± 0.063 & 0.667 ± 0.196 & 0.314 ± 0.219 & 0.687 ± 0.098 & \textbf{0.871 ± 0.093} & 0.233 ± 0.120 & 0.585 ± 0.054 & 0.690 ± 0.248 & 0.260 ± 0.181 \\
\hline
 & 1 & 0.528 ± 0.014 & 0.636 ± 0.382 & 0.138 ± 0.176 & 0.537 ± 0.205 & 0.799 ± 0.447 & \textbf{0.008 ± 0.019} & 0.514 ± 0.082 & 0.812 ± 0.251 & 0.100 ± 0.141 \\
\textbf{Nash-Met} & 3 & 0.538 ± 0.048 & 0.583 ± 0.381 & \textbf{0.043 ± 0.064} & 0.487 ± 0.173 & 0.483 ± 0.480 & 0.108 ± 0.181 & 0.550 ± 0.078 & 0.788 ± 0.216 & 0.180 ± 0.164 \\
 & 5 & 0.523 ± 0.074 & 0.770 ± 0.433 & 0.067 ± 0.066 & 0.575 ± 0.168 & 0.664 ± 0.385 & \textbf{0.050 ± 0.090} & 0.500 ± 0.050 & 0.368 ± 0.508 & \textbf{0.060 ± 0.089} \\
\hline
 & 1 & 0.559 ± 0.066 & 0.659 ± 0.383 & 0.090 ± 0.120 & 0.550 ± 0.204 & 0.763 ± 0.270 & 0.125 ± 0.128 & 0.592 ± 0.019 & 0.788 ± 0.093 & 0.120 ± 0.083 \\
\textbf{FEAST}& 3 & 0.579 ± 0.059 & 0.748 ± 0.271 & 0.157 ± 0.142 & 0.625 ± 0.139 & \textbf{0.899 ± 0.149} & \textbf{0.066 ± 0.095} & 0.600 ± 0.046 & 0.563 ± 0.363 & \textbf{0.099 ± 0.141} \\
 & 5 & 0.558 ± 0.077 & 0.742 ± 0.189 & 0.152 ± 0.100 & 0.575 ± 0.155 & 0.599 ± 0.547 & 0.075 ± 0.111 & 0.607 ± 0.056 & 0.645 ± 0.112 & 0.240 ± 0.219 \\
\hline
  & 1 & \textbf{0.615±0.000} &  0.837±0.012 & \textbf{0.060±0.051} & \textbf{0.781±0.044} &  0.792±0.059  &  0.062±0.088 & \textbf{0.643±0.152} &  \textbf{0.980±0.029}  & \textbf{0.000±0.000} \\
 \textbf{FairM2S} & 3 & \textbf{0.603±0.018} & 0.683±0.253  &  0.429±0.337 & \textbf{0.781±0.044} & 0.847±0.098 &  0.104±0.147 & \textbf{0.643±0.051} & \textbf{0.861±0.012} & 0.150±0.212\\
 & 5 & \textbf{0.603±0.127}  &  \textbf{0.914±0.121}  &  \textbf{0.012±0.017} & \textbf{0.750±0.088} & 0.833±0.236 & 0.062±0.088 & \textbf{0.679±0.202} & \textbf{0.900±0.007}& 0.100±0.000 \\
\hline
\bottomrule
\end{tabular}
}
\caption{Performance across datasets (SAVSD, StressID, AVD) and models. Best values per shot in bold: highest Accuracy, DI closest to 1, Eopp closest to 0.}
\label{tab:main_results_AV}
\end{table*}

\begin{figure}[t]
    \centering
    \includegraphics[scale=0.19]{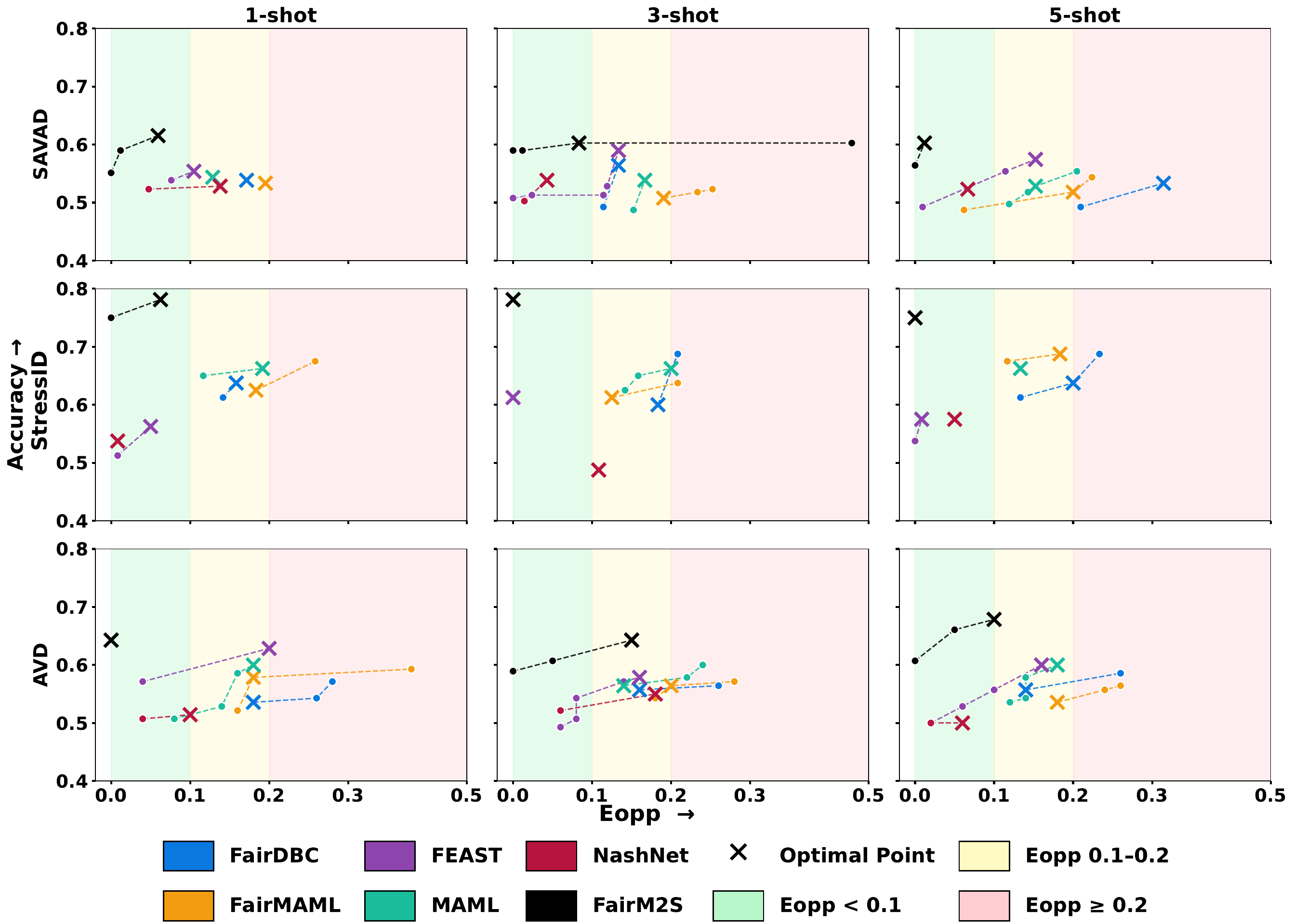}
    \caption{Pareto frontier comparison of models across datasets and shots. (\textbf{Best viewed in color})}
    \label{fig:pareto_grid_1}
\end{figure}

\begin{figure*}[t]
    \centering
    \includegraphics[width=\linewidth]{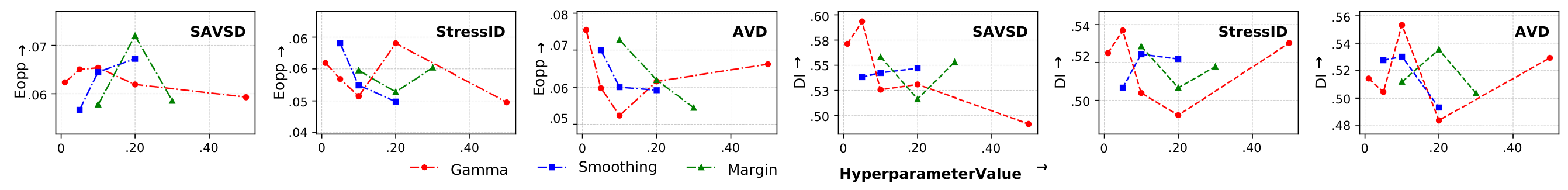}
    \caption{Fairness sensitivity of FairM2S across datasets and hyperparameters for Eopp and DI. (\textbf{Best viewed in color})}
    \label{fig:fairness_sensitivity}
\end{figure*}

\begin{figure}[t]
    \centering
    \includegraphics[width=\columnwidth]{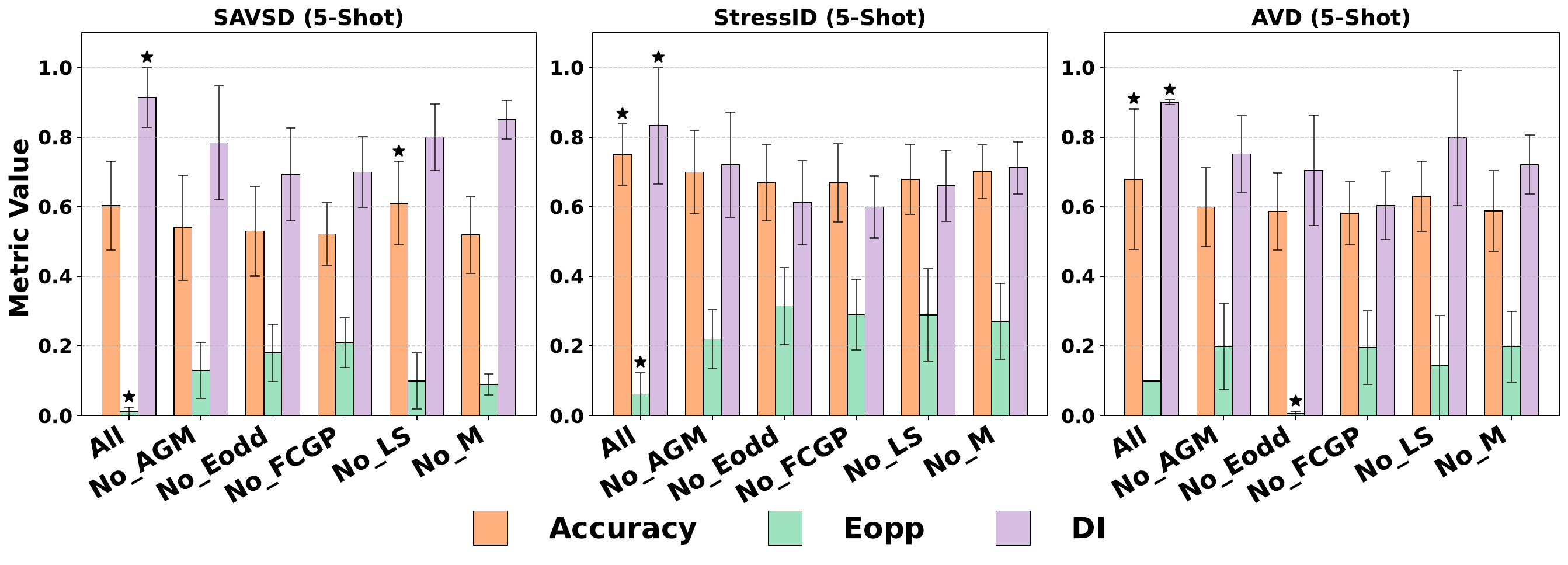}
    \caption{Ablation results at 5-Shot across datasets.}
    \label{fig:pareto_grid}
\end{figure}

\subsection{Performance Comparison}

As shown in Table~\ref{tab:main_results_AV}, FairM2S consistently outperforms baseline models across datasets and shot settings, achieving superior accuracy and fairness metrics. On SAVSD, FairM2S achieves the highest accuracy (0.615 at 1-shot, stable ~0.60 at 3 and 5-shots), reduces Eopp  by 80\% (0.060 to 0.012), and improves DI by 9\% (0.837 to 0.914). Furthermore, MAML achieves only 1.7\% accuracy but suffers a 58\% Eopp  increase (0.129 to 0.204) and 20\% DI decline (0.891 - 0.713), indicating fairness degradation. DBC F-MAML slightly improves Eopp but lags in accuracy, while Nash-Met has lowest Eopp at 3-shot (0.043) but low accuracy (0.538), showing a harsh trade-off. On StressID, FairM2S leads with high accuracy (0.78 at 1 and 3-shots) and moderate Eopp (0.062–0.104). FEAST nears FairM2S fairness (DI with 0.899) but trails 20\% in accuracy. MAML holds stable accuracy (0.66) but endures triple Eopp (0.19); Fair-MAML fluctuates in fairness and accuracy. Nash-Met gains fairness but at a significant accuracy loss. On AVD, FairM2S tops accuracy (0.643–0.679) and DI (0.86–0.98), with near-zero Eopp at 1-shot and slight rise (0.100) at 5-shot. Nash-Met and FEAST reach similar fairness but with up to 20\% accuracy drop.  Importantly, FairM2S uniquely improves or sustains accuracy and fairness as shots increase—a trend absent in baselines, which plateau or worsen fairness. On SAVSD, FairM2S reduces Eopp 80\% while MAML’s fairness declines 58\%. DI gains up to 9\% highlight superior group fairness scaling. Nash-Met and FEAST typify trade-offs, excelling in either metric but rarely both.

In sum, FairM2S consistently achieves superior accuracy and fairness, providing a robust solution for equitable stress detection in resource-limited settings.

\subsection{Fairness-Accuracy Tradeoff (Pareto Curve)}
The Pareto frontier analysis (Figure~\ref{fig:pareto_grid_1}) of accuracy versus Eopp across datasets and shot settings shows that \textbf{FairM2S} consistently achieves the best trade-off. On SAVSD, it attains high accuracy (e.g., $>0.60$ in 5-shot) with a minimal Eopp($0.012$), while MAML and FairMAML suffer from either lower accuracy or higher fairness gaps (Eopp $>$ 0.15). Under strict fairness constraints (Eopp $< 0.1$), baseline models drop below $0.58$ accuracy, but FairM2S maintains both high accuracy and fairness, defying the typical trade-off. As the number of shots increases, FairM2S’s curve shifts upward and leftward, indicating simultaneous improvements in accuracy and fairness—unlike other models, which tend to plateau or worsen. FEAST and NashNet sometimes approach FairM2S in either accuracy or fairness, but not both; on StressID, FEAST achieves Eopp $< 0.1$ but trails in accuracy by 13–20\%, while NashNet matches FairM2S in fairness at the cost of a 17–20\% drop in accuracy. DBC, FairMAML, and MAML consistently appear in the lower-right region of the plots, indicating poor trade-offs between accuracy and fairness. These results highlight that FairM2S uniquely expands the Pareto frontier, offering more balanced and superior solutions for fair stress detection.

\subsection{Fairness Sensitivity Across Hyperparameters}
Figure~\ref{fig:fairness_sensitivity} shows FairM2S’s sensitivity to key hyperparameters such as fairness weight \(\gamma\), smoothing factor \(\beta\), and margin loss weight \(\alpha\) across the SAVSD, StressID, and AVD datasets. The x-axis represents the hyperparameter values. The Eopp remains low (\(<0.08\)) and stable over broad ranges, indicating robust fairness unaffected by tuning. Eopp varies minimally (0.05–0.08) on SAVSD and StressID, with a slight increase on AVD at higher values. DI stays within a favorable range (0.50–0.60), reinforcing fairness stability. Unlike MAML, whose fairness metrics fluctuate with hyperparameter changes, FairM2S consistently maintains fairness and performance even as margin and smoothing increase, showing no trade-off or over-regularization risk.

\subsection{Ablation Study}
 
Figure~\ref{fig:pareto_grid} shows FairM2S (“All”) achieves the best balance of accuracy, Eopp, and DI, where $^\star$ marks the highest ACC, DI, and Eopp across datasets. Removing any fairness component—Adversarial Gradient Masking (No\_AGM), Equalized Odds Loss (No\_Eodd), Fairness-Constrained Gradient Projection (No\_FCGP), Label Smoothing (No\_LS), or Margin Loss (No\_M)—causes drops in accuracy or fairness. Omitting Eodd (No\_Eodd) or AGM (No\_AGM) results in the largest fairness gaps, while removing Margin Loss (No\_M) or Label Smoothing (No\_LS) mainly reduces accuracy. The full model achieves the highest accuracy showing that all components are essential for optimal, bias-resistant results.

\section{Conclusion and Future Work}

We propose FairM2S, a fairness-aware meta-learning framework that reduces gender bias while maintaining accuracy in few-shot multimodal stress detection. By incorporating fairness constraints such as adversarial gradient masking, fairness-constrained gradient projection, margin loss, and label smoothing, FairM2S effectively addresses biases in mental health AI. Evaluated on diverse datasets including SAVSD, it outperforms existing methods and enables more equitable stress detection. This work advances diagnostics for underserved groups and trustworthy AI. Future directions include integrating causal fairness, and validating on broader datasets to improve scalability and generalization.

\bibliography{aaai2026}
\clearpage
\end{document}